\documentclass[conference]{ieeeconf}
\usepackage[latin9]{inputenc}
\usepackage{textcomp}
\usepackage{amsmath}
\usepackage{amssymb}
\usepackage[unicode=true,
 bookmarks=true,bookmarksnumbered=true,bookmarksopen=true,bookmarksopenlevel=1,
 breaklinks=false,pdfborder={0 0 1},backref=false,colorlinks=false]
 {hyperref}
\hypersetup{pdftitle={Your Title},
 pdfauthor={Your Name},
 dvipdfmx,setpagesize=false,pdfborderstyle=,pdfpagelayout=OneColumn,pdfnewwindow=true,pdfstartview=XYZ,plainpages=false}

\makeatletter

\newcommand{\noun}[1]{\textsc{#1}}
\newcommand{\lyxmathsym}[1]{\ifmmode\begingroup\def\b@ld{bold}
  \text{\ifx\math@version\b@ld\bfseries\fi#1}\endgroup\else#1\fi}

\providecommand{\tabularnewline}{\\}

\IEEEoverridecommandlockouts
\usepackage{cite}
\usepackage{textcomp}
\usepackage{xcolor}
\usepackage[caption=false,font=footnotesize]{subfig}
\usepackage{resizegather}
\usepackage{import}
\usepackage{graphicx}

\makeatother

\begin{document}
\global\long\def\bbloss{L_{\text{bb}}}%

\global\long\def\locloss{L_{\text{loc}}}%

\global\long\def\confloss{L_{\text{conf}}}%

\global\long\def\bouloss{L_{\text{b}}}%

\global\long\def\diceloss{L_{\text{dice}}}%

\global\long\def\javecc{\boldsymbol{J}_{c}^{\lg}}%

\global\long\def\vece{\quat p_{e}^{\lg}}%

\global\long\def\dotvece{\dot{\quat p}_{e}^{\lg}}%

\global\long\def\veca{\quat p_{\la}^{\lg}}%

\global\long\def\dotveca{\dot{\quat p}_{\la}^{\lg}}%

\global\long\def\javeca{\boldsymbol{J}_{\la}^{\lg}}%

\global\long\def\vecasi{\quat p_{\si}^{\la}}%

\global\long\def\dotvecasi{\dot{\quat p}_{\si}^{\la}}%

\global\long\def\javecasi{\boldsymbol{J}_{\si}^{\la}}%

\global\long\def\vecsi{\quat t_{\si}^{\lg}}%

\global\long\def\dotvecsi{\dot{\quat t}_{\si}^{\lg}}%

\global\long\def\javecsi{\boldsymbol{J}_{\si}^{\lg}}%

\global\long\def\fworld{\mathcal{F}_{\text{W}}}%

\global\long\def\flg{\mathcal{F}_{\lg}}%

\global\long\def\fdrill{\mathcal{F}_{\text{R1}}}%

\global\long\def\centerp{\quat p_{c}}%

\global\long\def\edgep{\quat p_{e}}%

\global\long\def\sip{\quat t_{\si}}%

\global\long\def\lgp{\quat t_{\lg}}%

\global\long\def\dotsip{\dot{\quat t}_{\si}}%

\global\long\def\dotlgp{\dot{\quat t}_{\lg}}%

\global\long\def\dotqsi{\dot{\quat q}_{\si}}%

\global\long\def\dotqlg{\dot{\quat q}_{\lg}}%

\global\long\def\dotq{\dot{\quat q}}%

\global\long\def\qsi{\quat q_{\si}}%

\global\long\def\qlg{\quat q_{\lg}}%

\global\long\def\transjasi{\boldsymbol{J}_{t_{1}}}%

\global\long\def\transjalg{\boldsymbol{J}_{t_{2}}}%

\global\long\def\rotatejalg{\boldsymbol{J}_{r_{2}}}%

\global\long\def\cscope{\text{C1}}%

\global\long\def\cillu{\text{C2}}%

\global\long\def\disjascope{\mymatrix J_{d,\cscope}}%

\global\long\def\disjaillu{\mymatrix J_{d,\cillu}}%

\global\long\def\disjascopesafe{\mymatrix J_{d,\cscope,\text{safe}}}%

\global\long\def\disjaillusafe{\mymatrix J_{d,\cillu,\text{safe}}}%

\global\long\def\disscope{d_{\cscope}}%

\global\long\def\disillu{d_{\cillu}}%

\global\long\def\dotdisscope{\dot{d}_{\cscope}}%

\global\long\def\dotdisillu{\dot{d}_{\cillu}}%

\global\long\def\disscopesafe{d_{\cscope,\text{safe}}}%

\global\long\def\disillusafe{d_{\cillu,\text{safe}}}%

\global\long\def\dotdisscopesafe{\dot{d}_{\cscope,\text{safe}}}%

\global\long\def\dotdisillusafe{\dot{d}_{\cillu,\text{safe}}}%

\global\long\def\wsradius{r}%

\global\long\def\thetascope{\theta_{\cscope}}%

\global\long\def\thetascopesafe{\theta_{\cscope,\text{safe}}}%

\global\long\def\thetaillu{\theta_{\cillu}}%

\global\long\def\thetaillusafe{\theta_{\cillu,\text{safe}}}%

\global\long\def\o{\boldsymbol{0}}%

\global\long\def\lgdir{\quat l_{\lg}}%

\global\long\def\lgr{\quat r_{\lg}}%

\global\long\def\dotlgdir{\dot{\quat l}_{\lg}}%

\global\long\def\dotlgr{\dot{\quat r}_{\lg}}%

\global\long\def\shaftdis{d_{\text{shaft}}}%

\global\long\def\tipdis{d_{\text{tip}}}%

\global\long\def\quat#1{\boldsymbol{#1}}%

\global\long\def\dq#1{\underline{\boldsymbol{#1}}}%

\global\long\def\hp{\mathbb{H}_{p}}%

\global\long\def\dotmul#1#2{\left\langle #1,#2\right\rangle }%

\global\long\def\partialfrac#1#2{\frac{\partial\left(#1\right)}{\partial#2}}%

\global\long\def\totalderivative#1#2{\frac{d}{d#2}\left(#1\right)}%

\global\long\def\mymatrix#1{\boldsymbol{#1}}%

\global\long\def\vecthree#1{\operatorname{v}_{3}\left(#1\right)}%

\global\long\def\vecfour#1{\operatorname{v}_{4}\left(#1\right)}%

\global\long\def\haminuseight#1{\overset{-}{\mymatrix H}_{8}\left(#1\right)}%

\global\long\def\hapluseight#1{\overset{+}{\mymatrix H}_{8}\left(#1\right)}%

\global\long\def\haminus#1{\overset{-}{\mymatrix H}_{4}\left(#1\right)}%

\global\long\def\haplus#1{\overset{+}{\mymatrix H}_{4}\left(#1\right)}%

\global\long\def\norm#1{\left\Vert #1\right\Vert }%

\global\long\def\abs#1{\left|#1\right|}%

\global\long\def\conj#1{#1^{*}}%

\global\long\def\veceight#1{\operatorname{v}_{8}\left(#1\right)}%

\global\long\def\myvec#1{\boldsymbol{#1}}%

\global\long\def\imi{\hat{\imath}}%

\global\long\def\imj{\hat{\jmath}}%

\global\long\def\imk{\hat{k}}%

\global\long\def\dual{\varepsilon}%

\global\long\def\getp#1{\operatorname{\mathcal{P}}\left(#1\right)}%

\global\long\def\getpdot#1{\operatorname{\dot{\mathcal{P}}}\left(#1\right)}%

\global\long\def\getd#1{\operatorname{\mathcal{D}}\left(#1\right)}%

\global\long\def\getddot#1{\operatorname{\dot{\mathcal{D}}}\left(#1\right)}%

\global\long\def\real#1{\operatorname{\mathrm{Re}}\left(#1\right)}%

\global\long\def\imag#1{\operatorname{\mathrm{Im}}\left(#1\right)}%

\global\long\def\spin{\text{Spin}(3)}%

\global\long\def\spinr{\text{Spin}(3){\ltimes}\mathbb{R}^{3}}%

\global\long\def\distance#1#2#3{d_{#1,\mathrm{#2}}^{#3}}%

\global\long\def\distancejacobian#1#2#3{\boldsymbol{J}_{#1,#2}^{#3}}%

\global\long\def\distancegain#1#2#3{\eta_{#1,#2}^{#3}}%

\global\long\def\distanceerror#1#2#3{\tilde{d}_{#1,#2}^{#3}}%

\global\long\def\dotdistance#1#2#3{\dot{d}_{#1,#2}^{#3}}%

\global\long\def\distanceorigin#1{d_{#1}}%

\global\long\def\dotdistanceorigin#1{\dot{d}_{#1}}%

\global\long\def\squaredistance#1#2#3{D_{#1,#2}^{#3}}%

\global\long\def\dotsquaredistance#1#2#3{\dot{D}_{#1,#2}^{#3}}%

\global\long\def\squaredistanceerror#1#2#3{\tilde{D}_{#1,#2}^{#3}}%

\global\long\def\squaredistanceorigin#1{D_{#1}}%

\global\long\def\dotsquaredistanceorigin#1{\dot{D}_{#1}}%

\global\long\def\crossmatrix#1{\overline{\mymatrix S}\left(#1\right)}%

\global\long\def\constraint#1#2#3{\mathcal{C}_{\mathrm{#1},\mathrm{#2}}^{\mathrm{#3}}}%

\title{\textbf{Autonomous Robotic Drilling System for Mice Cranial Window
Creation: An Evaluation with an Egg Model}}
\author{Enduo~Zhao, Murilo~M.~Marinho, and Kanako~Harada\thanks{This
work was supported by JST Moonshot R\&D JPMJMS2033.}\thanks{(\emph{Corresponding
author:} Murilo~M.~Marinho)}\thanks{Enduo~Zhao, Murilo~M.~Marinho,
and Kanako~Harada are with the Department of Mechanical Engineering,
the University of Tokyo, Tokyo, Japan. \texttt{Emails:\{endowzhao1996,
murilo, kanakoharada,\}@g.ecc.u-tokyo.ac.jp}. }}
\maketitle
\begin{abstract}
Robotic assistance for experimental manipulation in the life sciences
is expected to enable precise manipulation of valuable samples, regardless
of the skill of the scientist. Experimental specimens in the life
sciences are subject to individual variability and deformation, and
therefore require autonomous robotic control. As an example, we are
studying the installation of a cranial window in a mouse. This operation
requires the removal of the skull, which is approximately 300 um thick,
to cut it into a circular shape 8 mm in diameter, but the shape of
the mouse skull varies depending on the strain of mouse, sex and week
of age. The thickness of the skull is not uniform, with some areas
being thin and others thicker. It is also difficult to ensure that
the skulls of the mice are kept in the same position for each operation.
It is not realistically possible to measure all these features and
pre-program a robotic trajectory for individual mice. The paper therefore
proposes an autonomous robotic drilling method. The proposed method
consists of drilling trajectory planning and image-based task completion
level recognition. The trajectory planning adjusts the z-position
of the drill according to the task completion level at each discrete
point, and forms the 3D drilling path via constrained cubic spline
interpolation while avoiding overshoot. The task completion level
recognition uses a DSSD-inspired deep learning model to estimate the
task completion level of each discrete point. Since an egg has similar
characteristics to a mouse skull in terms of shape, thickness and
mechanical properties, removing the egg shell without damaging the
membrane underneath was chosen as the simulation task. The proposed
method was evaluated using a 6-DOF robotic arm holding a drill and
achieved a success rate of 80\% out of 20 trials.
\end{abstract}

\section{Introduction}

A cranial window is a transparent observatory window created in the
skull of a mouse through which scientists can observe the mouse brain.
For example, human organoids are implanted into the mouse brain \cite{Koike2019}.
In this procedure, a microdrill is used to remove an 8 mm circular
patch of the mouse skull under a microscope, which is then replaced
with a cover glass. Any damage to the underlying brain will result
in a failed procedure.

\begin{figure}
\centering
\def\svgwidth{250pt}
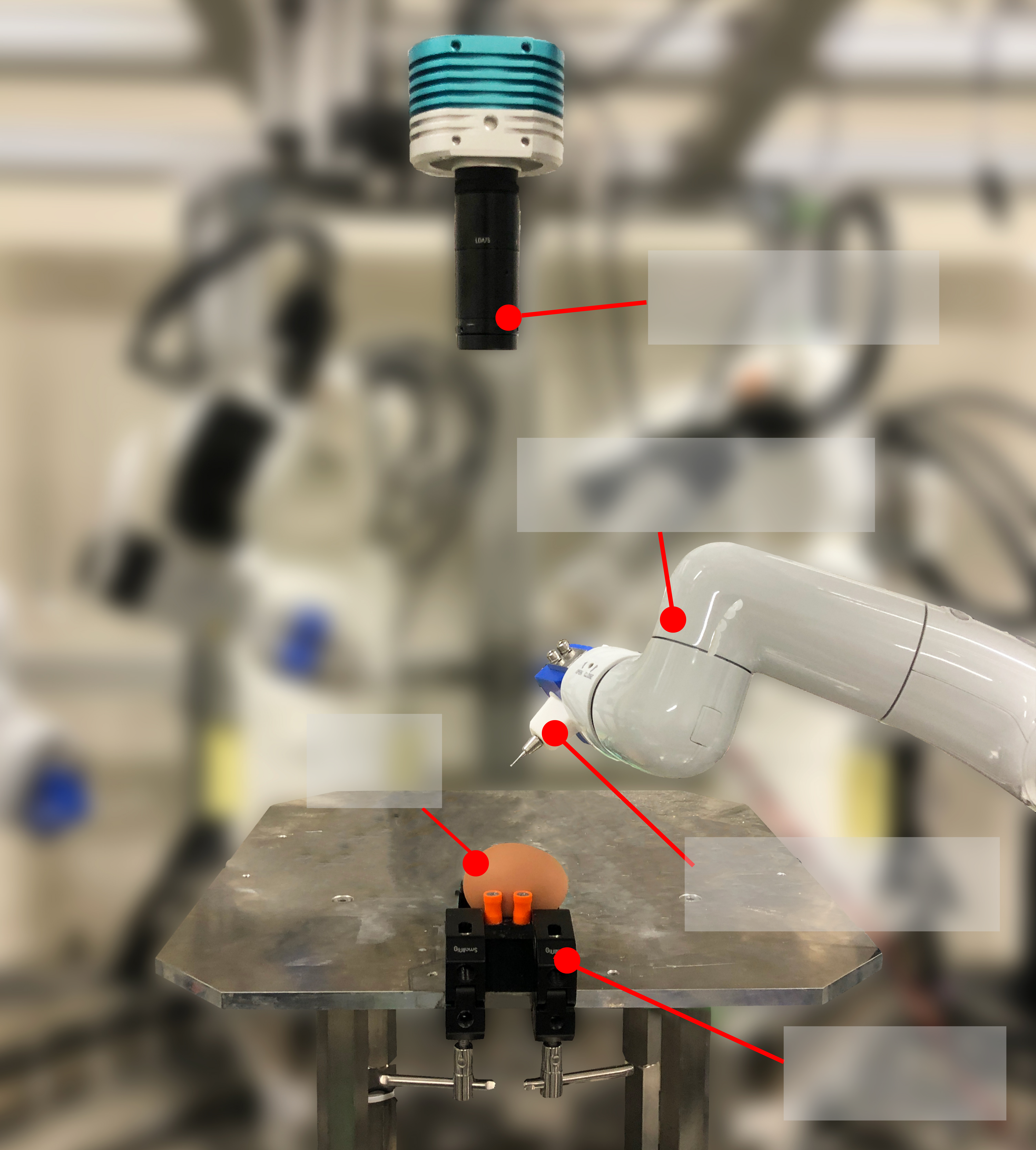

\caption{\label{fig:robot_system}The system setup used in this work, consisting
of one of the robotic arms of our AI-robot platform for scientific
exploration \cite{marques2022design}. For this work, we used the
arm with a micro drill as an end effector, a 4K camera, and a clamping
device for the egg.}
\end{figure}

Many devices and robotic systems have been developed to assist in
cranial window creation. Phuong \emph{et al.} \cite{Ly2020} developed
a robotic stereotaxic platform for small rodents by combining a skull
profiler sub-system and a six-degrees-of-freedom robotic platform.
Ghanbari \emph{et al. }\emph{\noun{\cite{Ghanbari2019} }}developed
a cranial microsurgery platform (Craniobot), based on a modified desktop
computer numerical controlled mill, which was used for precise microsurgical
procedures. Pak \emph{et al.} \cite{Pak2015} implemented a cranial
drilling robot which can judge whether the mouse cranium is penetrated
using conductance measurements to perform automated craniotomies.Our
group is exploring the concept of autonomously performing cranial
windows using a robotic manipulator retrofitted with a micro-drill
\cite{marques2022design}, as shown in Fig.~ \ref{fig:robot_system},
and cranial window installation is one of the target tasks to be demonstrated.

Obviously, it is not ethically acceptable to use mice to study autonomous
robot control at this stage of the research, and so we have chosen
to simulate the removal of the mouse skull without damaging the underlying
dura by removing a chicken egg shell without damaging the underlying
membrane. The characteristics of the egg shell are similar to those
of the mouse skull, and in fact egg shell drilling is used for training
in mouse surgery \cite{andreoli2018egg}.

\subsection{Related works}

In order to develop an autonomous robot drilling system, perception
is an important topic. There have been many works on the estimation
of drilling penetration and drilling status by different signals during
drilling procedures of human bones. Hu \emph{et al. }\cite{Hu2014}
, by analyzing the force signal during the screw-path-drilling process,
and Dai \emph{et al.} \cite{Dai2013}, by analyzing the electrical
impedance signal in bone-drilling process, succeeded to estimate the
relative position between the screw or drill and the bone being drilled
and the drilling penetration respectively. Ying \emph{et al.} \cite{Ying2020}
found that it is more effective to measure the force and sound signals
during the drilling process. With the help of neural network, they
achieved a high precision estimate of the progress of the drilling
procedure.

Considering the similarities and differences between drilling human
bones and mouse craniums, several works aimed to enhance the perception
of robot in drilling surgery for mouse cranium. Pohl \emph{et al.
}\cite{Pohl2011} designed a module for assisting the craniotomy for
mice, by measuring force and sound signals. The drilling penetration
was estimated and the drill feed speed is controlled. Pak \emph{et
al. }\cite{Pak2015} showed a penetration detection strategy using
a measurement circuit that detects electrical conductance between
the drill and the mouse. They reported that a sudden increase in the
electrical conductance could indicate when the skull was penetrated.

The aforementioned works can only estimate the drilling completion
progress at the point of contact between the drill and the skull,
because the signals of force, sound, and vibration are only generated
during contact. In this work, we aim to perceive the drilling completion
progress of the entire drilled region at once. To do so, we rely on
image processing using a neural network.Such network needs to serve
two functions. First, the drilled area must be detected. Second, the
pixel-wise completion level must be given. It is shown in \cite{Dvornik2017}
that the two tasks benefit from each other in terms of accuracy while
training.

Related to those goals, fast R-CNN and faster R-CNN \cite{Girshick2015,Ren2015}
can be used to obtain the target's bounding box using a CNN which
is then filtered using non-maximum suppression. Based on it, mask
R-CNN \cite{He2017} extends faster R-CNN by adding a branch for predicting
an object mask in parallel with the existing branch for bounding box
recognition. It offers high accuracy but has a comparatively slow
inference speed. Moreover, the popular object detector Single Shot
Detector (SSD) \cite{Liu2015} classifies all boxes directly using
a sliding window technique. To find items of varying sizes in a single
forward pass, SSD builds a scale pyramid. Deconvolutional Single Shot
Detector (DSSD) \cite{Fu2017} is a network created on the foundation
of SSD that combines a cutting-edge classifier (ResNet-101) with a
quick detection framework (SSD), and then adds deconvolution layers
to SSD+ResNet-101 to add additional large-scale context in object
detection and recognition.

Considering that the DSSD architecture consists of deconvolutional
layers, it is natural to try to do the semantic segmentation in the
same network because the purpose of a sequence of deconvolutional
layers is to increase the resolution of the output feature maps. Based
on this characteristic, other network architectures \cite{Simonyan2014,Noh2015,Kokkinos2016,Dvornik2017}
also aim to simultaneously perform semantic segmentation and object
detection. Our work is inspired by these previous attempts, where
the task of semantic segmentation is replaced by the completion level
map prediction.

\subsection{Statement of contributions\label{subsec:Statement-of-contributions}}

In this work, we develop an autonomous robotic drilling system with
only image feedback. For this, we (1) propose a trajectory planning
algorithm based on constrained splines that is adjusted in real time
by the image feedback; (2) propose a neural network to detect the
drilling area and obtain the pixel-wise completion level from the
microscopic image to update the planner trajectory; and (3) evaluated
our method in robotic drilling experiments using eggshells.

\begin{figure*}[t]
\centering
\def\svgwidth{500pt}
\fontsize{7pt}{7pt}\selectfont
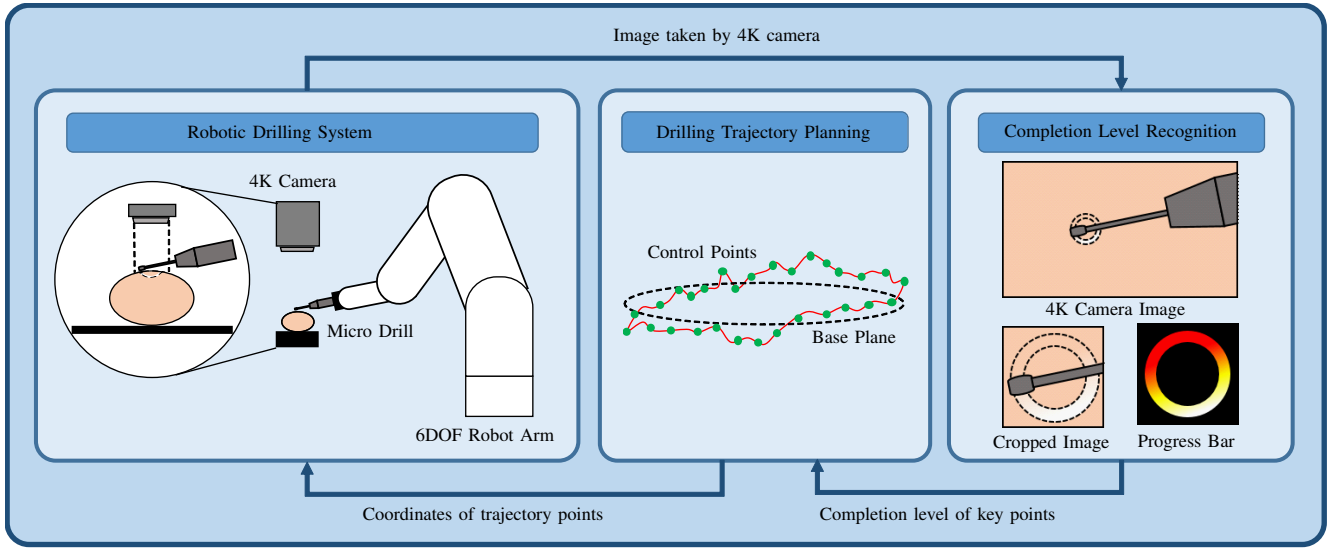

\caption{\label{fig:overall_system}Block diagram of the autonomous robotic
drilling system. The robotic system interacts with the egg and a 4K
camera provides images to the completion level recognition block.
The completion level recognition block outputs the depth of the points
for the trajectory planning block. Lastly, the trajectory planning
block uses the image information to update the trajectory and sends
setpoints to the robotic system, closing the loop.}
\end{figure*}

\section{Problem Statement\label{sec:Problem-Statement}}

Consider the setup shown in Fig.~\ref{fig:robot_system}, using one
of the robotic arms of our robot platform for scientific exploration
\cite{marques2022design}. Let $R1$ be the robotic arm (CVR038, Densowave,
Japan) with joint values $\quat q_{\text{R1}}\in\mathbb{R}^{6}$ holding
the micro drill (MD1200, Braintree Scientific, USA). The micro drill
is used to drill the egg fixed by a clamping device. Images are obtained
through a 4K microscope system (STC-HD853HDMI, Omron-Sentech, Japan)
from above.

\subsection{Goal}

Our goal is to autonomously drill the eggshell along a circular path
with respect to $\fdrill$ with the robot system using only image
as feedback.

\subsection{Premises}

Our robot controller was designed under the following premises:

\textbullet{} The joint velocities of the robot, $\dot{\quat q}\in\mathbb{R}^{6}$,
are defined by the output of the controller $\quat u_{\text{R1}}\in\mathbb{R}^{6}$
(Premise I).

\textbullet{} The solution $\quat u_{\text{R1}}=0_{\text{6}}$ must
always be feasible. That is, at worst the robots can stop moving (Premise
II).

\textbullet{} The robot has a kinematic model that is precise enough
for the task (Premise III).

\section{Proposed image-based autonomous drilling method}

The overview of the proposed method is shown in Fig.~\ref{fig:overall_system}.
In our strategy, the drill is autonomously controlled along a trajectory
planner (introduced in Section~\ref{subsec:Trajectory-planning-for})
in the $x\text{--}y$ plane while the $z-$axis coordinate of each
point is updated by the proposed image-based completion level recognition
system (introduced in Section~\ref{subsec:Completion-Level-Recognition}).

\subsection{\label{subsec:Trajectory-planning-for}Trajectory Planning}

Without loss of generality, let the goal be to plan a circular trajectory
in the $x\text{--}y$ plane with the depth modulated by the image-based
estimation algorithm.

We discretize the circular path into $n$ equal intervals resulting
in the trajectory points $\quat p_{i}\left(t\right)\triangleq\begin{bmatrix}p_{x,i} & p_{y,i} & p_{z,i}\left(t\right)\end{bmatrix}^{\text{T}}\in\mathbb{R}^{3}$,
$i=1,\cdots,n\in\mathbb{N}$. In addition, let $p_{x,i},p_{y,i}$
be constant while only $p_{z,i}\left(t\right)$ varies on time. At
$t=0$, all the points have the same $z-$coordinate value $p_{z,i}\left(0\right)=p_{z}\left(0\right)~\forall~i$.
We modulate the velocity by which each trajectory point goes down,
$v_{z,i}\left(t\right)=\dot{p}_{z,i}\left(t\right)$, using the completion
level of each point estimated by the image.

With this, we expect to autonomously drill a circular patch without
making further assumptions on the drilling surface. Then, we interpolate
between the $n$ trajectory points to obtain a smooth trajectory using
a constrained spline interpolation algorithm.

\subsubsection{Z-coordinate Value Calculation}

Our proposal is to modulate the lowering velocity of each trajectory
point as
\begin{align}
v_{z,i}\left(t\right)= & \left(1-c_{i}\right)v_{z}\left(0\right),\label{eq:vp}
\end{align}
where $v_{z}\left(0\right)$ is the initial downward velocity and
$0\leqslant c_{i}\leqslant1$ is the drilling completion level at
point $i$.

The position is then calculated through simple integration with a
sampling time $T$ as
\begin{align}
p_{z,i}\left(t+T\right)= & p_{z,i}\left(t\right)-v_{z,i}\left(t\right)T.\label{eq:zp}
\end{align}

The completion level map gives us enough information to assume the
completion level of $n$ points. We interpolate between them with
a constrained cubic spline to obtain a smooth trajectory.

\subsubsection{Constrained Cubic Spline Interpolation}

\begin{figure}
\centering
\fontsize{5pt}{5pt}\selectfont
\def\svgwidth{250pt}
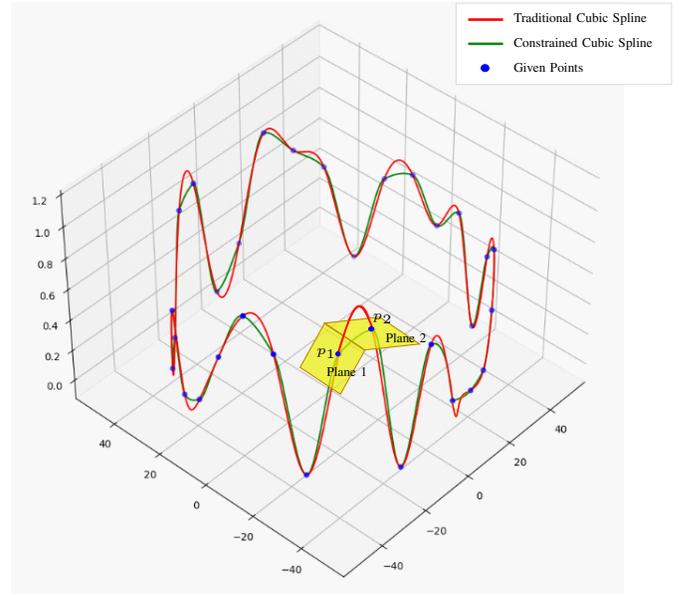

\caption{\label{fig:curve}Generated curve with $n=30$ points along the circular
path. The blue points are the sampling points, the red line is the
result of traditional spline interpolation, and the green line is
the result of constrained spline interpolation. We can see the red
line overshoots while the green does not.}
\end{figure}

Cubic spline interpolation has been widely applied for path planning
but it is, by itself, unsuitable for our application as it might overshoot
between two trajectory points and cause rupture to the membrane (see
Fig.~\ref{fig:curve}). Instead, we propose the use of constrained
spline interpolation by eliminating the requirement for equal second
order derivatives at every point and replacing it with specified first
order derivatives \cite{kruger2003constrained}.

Consider any two contiguous trajectory points $\quat p_{i}\triangleq\begin{bmatrix}p_{x,i} & p_{y,i} & p_{z,i}\end{bmatrix}^{\text{T}}$
and $\quat p_{i+1}\triangleq\begin{bmatrix}p_{x,i+1} & p_{y,i+1} & p_{z,i+1}\end{bmatrix}^{\text{T}}$,
$i=1,\cdots,n\in\mathbb{N}$. Let $u$ be a free parameter to generate
a third-degree polynomial curve $\quat s_{i}\left(u\right)$ between
these two points. Let $u=0$ be the beginning of the curve, $\quat p_{i}$,
and $u=1$ be the end of the curve, $\quat p_{i+1}$. Notice that
when $i=n$, $\quat p_{n}$ is the last point on the trajectory so
$\quat s_{n}\left(u\right)$ will be the curve between the last point
$\quat p_{n}$ and the first point $\quat p_{1}$. Then $0\leqslant u\leqslant1$
are defined by
\begin{alignat}{1}
\quat s_{i}\left(u\right) & =\begin{bmatrix}s_{x,i}\left(u\right) & s_{y,i}\left(u\right) & s_{z,i}\left(u\right)\end{bmatrix}^{\text{T}}\nonumber \\
 & =\quat a_{i}^{\text{T}}+\quat b_{i}^{\text{T}}u+\quat c_{i}^{\text{T}}u^{2}+\quat d_{i}^{\text{T}}u^{3},\label{eq:cons1}
\end{alignat}

where
\[
\begin{array}{c}
\quat a_{i}^{\text{T}}=\begin{bmatrix}a_{x,i} & a_{y,i} & a_{z,i}\end{bmatrix}^{\text{T}}\\
\quat b_{i}^{\text{T}}=\begin{bmatrix}b_{x,i} & b_{y,i} & b_{z,i}\end{bmatrix}^{\text{T}}\\
\quat c_{i}^{\text{T}}=\begin{bmatrix}c_{x,i} & c_{y,i} & c_{z,i}\end{bmatrix}^{\text{T}}\\
\quat d_{i}^{\text{T}}=\begin{bmatrix}d_{x,i} & d_{y,i} & d_{z,i}\end{bmatrix}^{\text{T}}
\end{array}.
\]
The first order derivative and second order derivative of the curve
are
\begin{equation}
\quat s_{i}^{\prime}\left(u\right)=\begin{bmatrix}s_{x,i}^{\prime} & s_{y,i}^{\prime} & s_{z,i}^{\prime}\left(u\right)\end{bmatrix}^{\text{T}}=\quat b_{i}^{\text{T}}+2\quat c_{i}^{\text{T}}u+3\quat d_{i}^{\text{T}}u^{2}\label{eq:cons2}
\end{equation}
and
\begin{equation}
\quat s_{i}^{\prime\prime}\left(u\right)=\begin{bmatrix}s_{x,i}^{\prime\prime} & s_{y,i}^{\prime\prime} & s_{z,i}^{\prime\prime}\left(u\right)\end{bmatrix}^{\text{T}}=2\quat c_{i}^{\text{T}}+6\quat d_{i}^{\text{T}}u.\label{eq:cons3}
\end{equation}

In order to calculate the parameters of the curve, we have three conditions.
First, the curve must pass through points $\quat p_{i}$ and $\quat p_{i+1}$,
that is,
\begin{equation}
\begin{array}{c}
\quat s_{i}\left(0\right)=\begin{bmatrix}p_{x,i} & p_{y,i} & p_{z,i}\end{bmatrix}^{\text{T}}\\
\quat s_{i}\left(1\right)=\begin{bmatrix}p_{x,i+1} & p_{y,i+1} & p_{z,i+1}\end{bmatrix}^{\text{T}}
\end{array}.\label{eq:cons4}
\end{equation}
Second, the first order derivative of the whole trajectory must be
continuous, which means the first order derivatives must be the same
at each interval end, that is,
\begin{equation}
\quat s_{i}^{\prime}\left(1\right)=\quat s_{i+1}^{\prime}\left(0\right).\label{eq:const5}
\end{equation}

The third conditions for traditional traditional cubic spline interpolation
and for constrained cubic spline interpolation are different.

For traditional cubic spline interpolation, the second order derivative
of the whole trajectory must be continuous, which means the second
order derivative on the left side of a point, $\quat s_{i}^{\prime\prime}\left(1\right)$,
must be the same as that on the right side, $\quat s_{i+1}^{\prime\prime}\left(0\right)$.
As a result, the generated trajectory is smooth enough but overshoot
might occur.

For constrained cubic spline interpolation, we aim to solve the overshoot
problem by sacrificing the smoothness so that the second order derivative
doesn't need to be continuous. Instead, the value of first order derivative
at each point is known, that is,
\begin{equation}
\begin{array}{c}
\quat s_{i}^{\prime}\left(0\right)=\begin{bmatrix}p_{x,i}^{\prime} & p_{y,i}^{\prime} & p_{z,i}^{\prime}\end{bmatrix}^{\text{T}}\\
\quat s_{i}^{\prime}\left(1\right)=\begin{bmatrix}p_{x,i+1}^{\prime} & p_{y,i+1}^{\prime} & p_{z,i+1}^{\prime}\end{bmatrix}^{\text{T}}
\end{array}.\label{eq:cons7}
\end{equation}
By combining \eqref{eq:cons1}, \eqref{eq:cons2}, \eqref{eq:cons4},
\eqref{eq:const5} and \ref{eq:cons7}, we get
\[
\quat a_{i}^{\text{T}}=\left[\begin{array}{c}
p_{x,i}\\
p_{y,i}\\
p_{z,i}
\end{array}\right],\,\quat b_{i}^{\text{T}}=\left[\begin{array}{c}
p_{x,i}^{\prime}\\
p_{y,i}^{\prime}\\
p_{z,i}^{\prime}
\end{array}\right],
\]

\[
\quat c_{i}^{\text{T}}=\left[\begin{array}{c}
3\left(p_{x,i+1}-p_{x,i}\right)-\left(p_{x,i+1}^{\prime}+2p_{x,i}^{\prime}\right)\\
3\left(p_{y,i+1}-p_{y,i}\right)-\left(p_{y,i+1}^{\prime}+2p_{y,i}^{\prime}\right)\\
3\left(p_{z,i+1}-p_{z,i}\right)-\left(p_{z,i+1}^{\prime}+2p_{z,i}^{\prime}\right)
\end{array}\right],
\]

\[
\quat d_{i}^{\text{T}}=\left[\begin{array}{c}
2\left(p_{x,i}-p_{x,i+1}\right)+p_{x,i+1}^{\prime}+p_{x,i}^{\prime}\\
2\left(p_{y,i}-p_{y,i+1}\right)+p_{y,i+1}^{\prime}+p_{y,i}^{\prime}\\
2\left(p_{z,i}-p_{z,i+1}\right)+p_{z,i+1}^{\prime}+p_{z,i}^{\prime}
\end{array}\right].
\]

With the constrained cubic spline interpolation described above, we
generate a curve as depicted in green in Fig.~\ref{fig:curve} and
compare it with a traditional cubic spline interpolation in red. We
can see in Fig.~\ref{fig:curve} that the green curve is enveloped
by the tangent planes but red curve is not, hence solving the problem
of overshoot.

\subsection{Completion Level Recognition\label{subsec:Completion-Level-Recognition}}

In the completion level recognition system, we aim to detect the drilling
area and predict the drilling completion level simultaneously using
a single neural network. It is known that detection and semantic segmentation
benefit from each other in training.

\subsubsection{Network architecture}

\begin{figure*}
\centering
\fontsize{5pt}{5pt}\selectfont
\def\svgwidth{500pt}
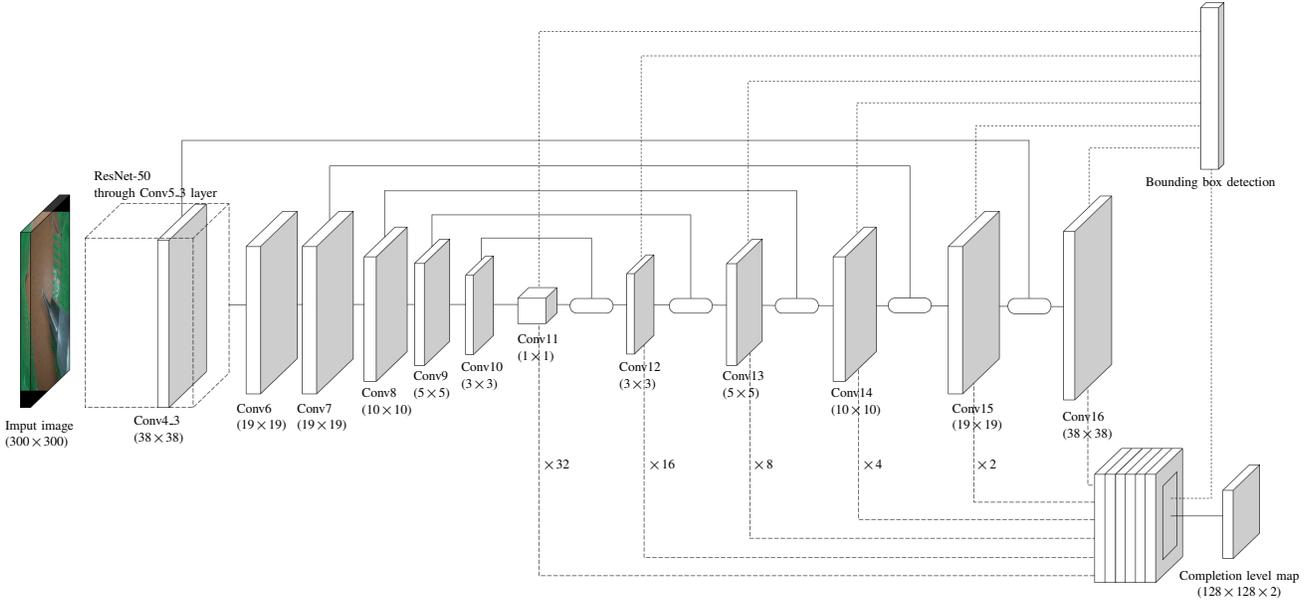

\caption{\label{fig:overall_network_structure}The overall network architecture
for the drilling area detection and completion level prediction. On
the left, ResNet-50 is applied as a feature extractor. It is followed
by the downscale-stream and the upscale-stream, which consist of a
sequence of deconvolution layers interleaved with conv-skip connection.
The localization and classification of bounding boxes (top) and pixelwise
completion level map (bottom) are performed in a multiscale fashion
by single convolutional layers operating on the output of deconvolution
layers.}
\end{figure*}

The overview of the architecture is shown in Fig.~\ref{fig:overall_network_structure}.
The structure of the network is inspired by DSSD \cite{Fu2017}, which
is a object detection network with an encoder-decoder structure that
provides a reasonable trade-off between accuracy and speed. One notable
difference is that we crop and concentrate the activation of the downscale
stream and the upscale stream together as a completion level prediction
branch.

\paragraph{Backbone of the network}

The input image is resized to ($300\times300$) and processed with
a convolutional neural network in order to obtain a map that carries
high-level features. ResNet-50 is applied as a feature extractor.

The feature extractor network ResNet-50 is followed by a series of
convolutional, pooling layers, and convolutional layers.

\paragraph{Bounding-box detection branch}

For bounding-box detection, we apply a similar approach to DSSD which
adds several residual structures before using multi-scale feature
maps for classification and regression. A single conv-skip connection
is applied to process the output of feature layers and obtain the
bounding box of the drilling area.

\paragraph{Completion prediction branch}

In our completion prediction branch, the desired output is a ($128\times128\times2$)
tensor where $128\times128$ is the size of the output picture and
2 is the number of channels. In order to use the feature information
of each layer as much as possible, we upsample the output of conv11,
conv12, conv13, conv14 and conv15 all into the size of conv16, then
concentrate them together and follow that by a convolutional layer
to change the number of the channel into 2. Finally, we crop the 2-channel
output according to the relative position of the bounding box and
the overall input image and resize it into ($128\times128\times2$).
One channel refers to the drill channel and another one is the completion
level map, whose grayscale $(0\lyxmathsym{\textendash}1)$ of each
pixel refers to the percentage of completion for this pixel.

\subsubsection{Training Dataset}

Generally, ground truth images with full manual annotation are required
for multi-task learning. In this work, we expect our network to output
a bounding box of the drilling area and a completion level map. So
it would be necessary to manually annotate the coordinates of the
four corners of the bounding box for the drilling area and a cropped
completion level map within the drilling area for the completion level
prediction branch as ground truth.

To generate the completion level maps, one approach would be to manually
annotate each pixel manually with a grayscale value ranging from 0
to 100. Considering the complexity of this approach, we propose a
more treatable annotation strategy.

Given an image of eggshell drilling, first we define 6 classes for
a semantic segmentation classifier: drill, $0\,\%$ drilled (background),
$25\,\mathrm{\%}$ penetrated, $50\,\%$ penetrated, $75\,\%$ penetrated,
and $100\,\%$ penetrated. The percentage of penetration is defined
subjectively with our experience in pilot studies. Second, in a higher
level in the hierarchy, we separate the image into confidence maps.
The first image is the confidence map for pixels of the drill, with
255 corresponding to the full confidence, so that we can filter it
out. The second confidence map corresponds to the completion level
map, where 0 corresponds to undrilled pixels and 100 corresponds to
completely drilled pixels. . Lastly, taking advantage of the expected
continuity in drilled regions, a Gaussian filter is applied to smooth
the completion level map channel so that its greyscale values of pixels
are continuous from 0 to 100.

A 4-dimensional vector ($x_{1},\,y_{1},\,x_{2},\,y_{2}$) is used
as the ground truth of the bounding box detection branch, which is
generated from the completion level map channel by noting the minimum
value and maximum value from width and height direction of those pixels
whose grayscale values are not 0. After that, we crop both the completion
level map channel and drill channel using the coordinate of the bounding
box ($x_{1},\,y_{1}$) and ($x_{2},\,y_{2}$) and then resize them
into $128\times128$. Merging two channels together we can get a $128\times128\times2$
matrix as the ground truth of the completion level prediction branch.

At the training stage, we used common image augmentation methods:
random flip, rotate, random crop, and random change of brightness
and contrast. We manually labeled 518 images and 16,576 after augmentation.
In total, 13,260 images were used for training, 1,658 were used for
validation and 1,658 were used for testing.

\begin{figure*}
\centering
\def\svgwidth{500pt}
%% Creator: Inkscape 1.2.1 (9c6d41e410, 2022-07-14), www.inkscape.org
%% PDF/EPS/PS + LaTeX output extension by Johan Engelen, 2010
%% Accompanies image file 'fig8_.pdf' (pdf, eps, ps)
%%
%% To include the image in your LaTeX document, write
%%   \input{<filename>.pdf_tex}
%%  instead of
%%   \includegraphics{<filename>.pdf}
%% To scale the image, write
%%   \def\svgwidth{<desired width>}
%%   \input{<filename>.pdf_tex}
%%  instead of
%%   \includegraphics[width=<desired width>]{<filename>.pdf}
%%
%% Images with a different path to the parent latex file can
%% be accessed with the `import' package (which may need to be
%% installed) using
%%   \usepackage{import}
%% in the preamble, and then including the image with
%%   \import{<path to file>}{<filename>.pdf_tex}
%% Alternatively, one can specify
%%   \graphicspath{{<path to file>/}}
%% 
%% For more information, please see info/svg-inkscape on CTAN:
%%   http://tug.ctan.org/tex-archive/info/svg-inkscape
%%
\begingroup%
  \makeatletter%
  \providecommand\color[2][]{%
    \errmessage{(Inkscape) Color is used for the text in Inkscape, but the package 'color.sty' is not loaded}%
    \renewcommand\color[2][]{}%
  }%
  \providecommand\transparent[1]{%
    \errmessage{(Inkscape) Transparency is used (non-zero) for the text in Inkscape, but the package 'transparent.sty' is not loaded}%
    \renewcommand\transparent[1]{}%
  }%
  \providecommand\rotatebox[2]{#2}%
  \newcommand*\fsize{\dimexpr\f@size pt\relax}%
  \newcommand*\lineheight[1]{\fontsize{\fsize}{#1\fsize}\selectfont}%
  \ifx\svgwidth\undefined%
    \setlength{\unitlength}{3022.1187946bp}%
    \ifx\svgscale\undefined%
      \relax%
    \else%
      \setlength{\unitlength}{\unitlength * \real{\svgscale}}%
    \fi%
  \else%
    \setlength{\unitlength}{\svgwidth}%
  \fi%
  \global\let\svgwidth\undefined%
  \global\let\svgscale\undefined%
  \makeatother%
  \begin{picture}(1,0.22244169)%
    \lineheight{1}%
    \setlength\tabcolsep{0pt}%
    \put(0,0){\includegraphics[width=\unitlength,page=1]{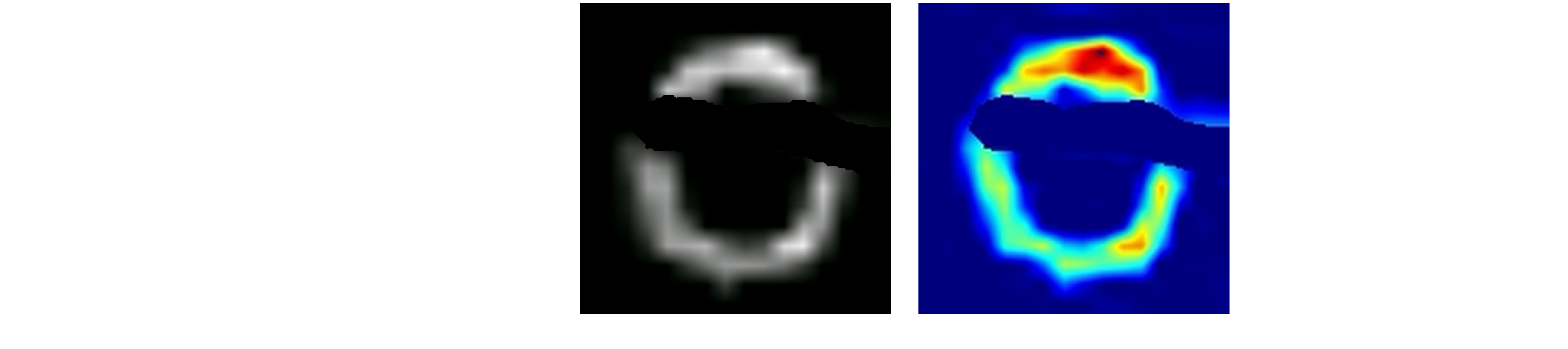}}%
    \put(0.68204686,0.0039586){\color[rgb]{0,0,0}\makebox(0,0)[t]{\lineheight{1.25}\smash{\begin{tabular}[t]{c}(c)\end{tabular}}}}%
    \put(0.17645086,0.00341719){\color[rgb]{0,0,0}\makebox(0,0)[t]{\lineheight{1.25}\smash{\begin{tabular}[t]{c}(a)\end{tabular}}}}%
    \put(0.46777172,0.00341719){\color[rgb]{0,0,0}\makebox(0,0)[t]{\lineheight{1.25}\smash{\begin{tabular}[t]{c}(b)\end{tabular}}}}%
    \put(0,0){\includegraphics[width=\unitlength,page=2]{fig8_.pdf}}%
    \put(0.90073181,0.00373241){\color[rgb]{0,0,0}\makebox(0,0)[t]{\lineheight{1.25}\smash{\begin{tabular}[t]{c}(d)\end{tabular}}}}%
  \end{picture}%
\endgroup%

\caption{\label{fig:result}Output image of the network. (a) Result of bounding
box detection; (b) The completion level map; (c) The heatmap of completion
level map for visualization; (d) The progress bar for visualization.}
\end{figure*}

\subsubsection{Network training}

\paragraph{Loss function}

The loss function of our training procedure contains three parts:
the loss function of bounding box detection $\bbloss$ for the drilling
area, the loss function of the drill semantic segmentation $L_{\text{drill}}$
for the drill, and the loss function of completion level prediction
$L_{\text{com}}$.

$\bbloss$ is a weighted sum of the localization loss $\locloss$
and the confidence loss $\confloss$, given by
\begin{align}
L_{\text{bb}}= & \frac{1}{N}\left(L_{\text{loc}}+L_{\text{conf}}\right).\label{eq:Lbb}
\end{align}
where $N$ is the number of matched default boxes.If the number of
matched default boxes is 0, we set $N$ as 1.

$L_{\text{drill}}$ is a weighted sum of dice loss $L_{\text{dice}}$
and boundary loss $L_{\text{b}}$, given by
\begin{align}
L_{\text{drill}} & =\text{\ensuremath{\alpha L_{\text{dice}}}}+\left(1-\alpha\right)L_{\text{b}}.\label{eq:Ldrill}
\end{align}
where $\alpha=max\left(0.01,\,1-0.01\times epoch\right)$. We can
see that as the training process progresses, the weight of $L_{\text{b}}$
continuously increases.

$L_{\text{com}}$ is L2 loss function which calculates the addition
of the absolute value of the difference of each according pixel.

The overall loss function $L$ is given by
\begin{align}
L= & L_{\text{bb}}+L_{\text{drill}}+L_{\text{com}}.\label{eq:L}
\end{align}

We could not find noticeable improvements in preliminary experiments
in which we tested different weights for the summation of loss functions.

\paragraph{Training condition}

During the training stage, the images are trained in a total of 6,000
epochs with the Adam optimizer. The training rate is set as $0.01\times0.9995^{n}$
while $n$ is the number of epochs, which is changeable with the training
process. Moreover, we used a batch size of 16 during the training
stage. All the architectures were implemented in Python 3.8 using
PyTorch 1.10.1 and CUDA 11.3 and executed in Ubuntu 20.04 with an
NVIDIA RTX A5000 graphics card.

\subsubsection{Training Result and Evaluation}

An example of output results is shown in Fig.~\ref{fig:result}.
The network is able to output a bounding box for drilling area (Fig.~\ref{fig:result}-(a))
and the completion level map for the area (Fig.~\ref{fig:result}-(b)).
The completion levels of parts that are drilled and not occluded
by the drill are shown in the completion level map, while undrilled
or occluded parts are suppressed. We can visualize the completion
level map using heatmap (Fig.~\ref{fig:result}-(c)) where color
is red if the completion level is higher and the color is blue if
the completion level is blue. By adding and normalizing the according
pixel value, we can easily calculate a value between 0 and 1 for each
angle along a circle and generate a drilling progress bar (Fig.~\ref{fig:result}-(d))
based on the completion level map to show the drilling progress. It
is important to note that the progress bar is generated by not only
the completion level map in this moment but also those in the past
because we expect the drilling progress of an area to be monotonically
increasing. By setting $n$ discrete points evenly along the circle
on the progress bar, the value $c_{i}$ of point $i$, which is the
completion level at point $i$, is used to calculate the lowering
velocity of drill by \eqref{eq:vp}.

mAP (mean Average Precision) is applied as the evaluation metric of
object detection task while MAPE (Mean Absolute Percentage Error)
is applied as the evaluation metric of completion level prediction
task. Our network architecture is able to reach up 78.5 in mAP for
detection task and $15.05\,\%$ in MAPE for prediction task while
the speed is $72\,\mathrm{Hz}$. We expected the result to be $\text{mAP}\,>\,75$,
$\text{MAPE}\,<\,20\,\%$ and real-time ($\mathrm{fps\,}>\,60$),
so our work is able to achieve the acquirement and have a good trade-off.

\begin{table}
\centering{}\caption{\label{tab:parameter-choice}The selected parameters during the experiment.}
\begin{tabular}{cccc}
\hline 
Parameters & $v_{\text{0}}\,(\text{m/s})$ & $R\,(\text{mm})$ & $f\,(\text{Hz})$\tabularnewline
\hline 
Selected value & \textbf{$6\times10^{-6}$} & 8 & 30\tabularnewline
\hline 
\end{tabular}
\end{table}

\section{Experiment}

An experiment was conducted to evaluate the feasibility of the autonomous
robotic drilling system.

The system is set up as shown in Fig.~\ref{fig:robot_system}.
An egg is fixed stably using a clamping device. The micro drill is
held by a robot arm and a 4K camera is set vertically above the center
of the egg to observe the drilling procedure. An air compressor with
a silicone tube is applied to blow away the shell dust emitted from
drilling the surface of the eggshell. Communication with the robot
was enabled by the SmartArmStack\footnote{https://github.com/SmartArmStack}.
The quaternion algebra and robot kinematics were implemented using
DQ Robotics \cite{adorno2021dqrobotics} with Python3.

Before the experiment starts, the drill is teleoperated by an operator
so that it is within the field-of-view of the camera throughout the
entire circular path. We planned the drilling trajectory with the
method mentioned in Sec.~\ref{subsec:Trajectory-planning-for}. 

We performed a total of 20 trials with 20 different eggs of random
shape, size and thickness of shell. After the autonomous drilling
algorithm stopped, we tried to remove the resected circular shell
piece manually using tweezers. If the eggshell could be removed without
damaging the membrane beneath the eggshell, the experiment was deemed
successful. On the other hand, if the membrane broke during the drilling
or removal procedure, or the eggshell was not easily removed, the
experiment was counted as a failure. If the robot over-drilled the
egg and ruptured the membrane, this would also be considered a failure.

\subsection{Preparation}

The parameters used in the experiment, namely the the initial speed
$v_{\text{0}}$ of the drill in \eqref{eq:vp}, the radius of the
circular drilling path $R$, and the frequency of the transmission
of the calculated $z-$positions to the robot $f$ are summarized
in Table~\ref{tab:parameter-choice}. Also, the drilling will stop
autonomously when the system judges the drilling procedure as complete.
This happens when $80\,\%$ of points on the drilling path achieve
at least $85\,\%$ of completion level.

\subsection{Results and discussion}

\begin{figure}
\centering
\def\svgwidth{250pt}
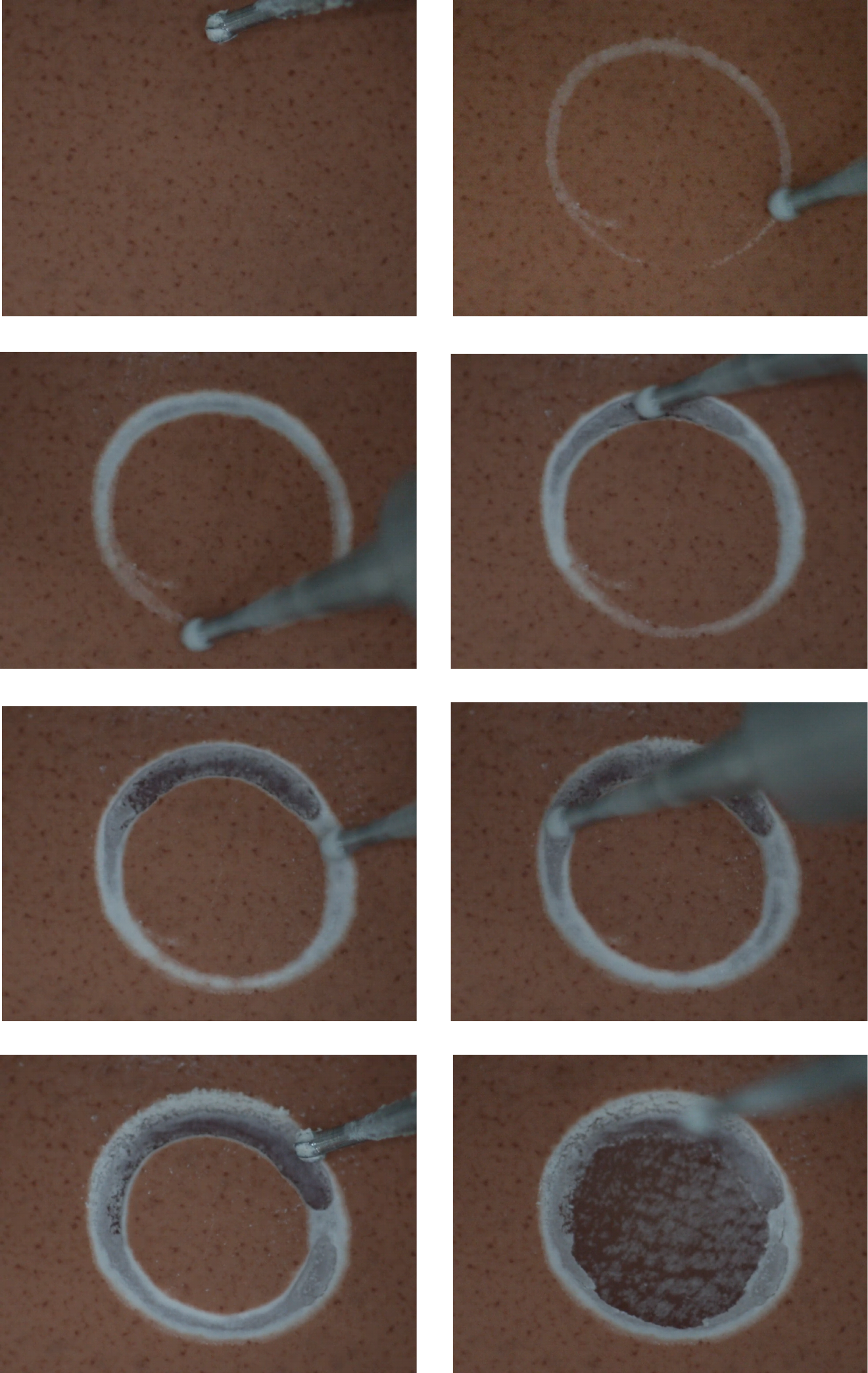

\caption{\label{fig:drillingsnapshots}A series of snapshots of a case of success
from the start of drilling to the removal of resected circular shell
piece.}
\end{figure}

In the experiment, 16 out of 20 cases were successful, resulting in
a success ratio of $80\,\%$. Fig. \ref{fig:drillingsnapshots} shows
a typical case of success where the drilling procedure will stop automatically
when it is judged as complete and the circular patch can be removed
perfectly without the membrane broke. The result can initially prove
the feasibility of the autonomous robotic drilling scheme.

The average required time for successful drilling was $16.8\,\text{min}$,
which is acceptable but we still expect it to be shorter considering
the drilling time of 7 out of 16 cases of success is over $25\,\text{min}$.
The main reason for the excessive drilling time could be a tilted
initial pose of the egg. Considering the drill trajectory in the $x\text{--}y$
plane is a circle and the $z$-coordinate of any point on the trajectory
at the initial moment are the same, if the initial pose of the egg
is tilted, the distance of any point on the drilling trajectory from
the initial position to its projection on the outside of the eggshell
differ much. As a result, the drill touches the eggshell at one point
on the drilling trajectory sooner than another, even already reach
$100\,\%$ completion level at one point and doesn't touch the eggshell
at another, resulting in a long drilling time. On the other hand,
if the initial pose of the egg is not tilted, the drill touches any
point of the circular trajectory almost at the same time and finishes
drilling almost simultaneously, leading to a short drilling time.

\begin{figure}
\centering
\def\svgwidth{250pt}
%% Creator: Inkscape 1.2.1 (9c6d41e410, 2022-07-14), www.inkscape.org
%% PDF/EPS/PS + LaTeX output extension by Johan Engelen, 2010
%% Accompanies image file 'fig17.pdf' (pdf, eps, ps)
%%
%% To include the image in your LaTeX document, write
%%   \input{<filename>.pdf_tex}
%%  instead of
%%   \includegraphics{<filename>.pdf}
%% To scale the image, write
%%   \def\svgwidth{<desired width>}
%%   \input{<filename>.pdf_tex}
%%  instead of
%%   \includegraphics[width=<desired width>]{<filename>.pdf}
%%
%% Images with a different path to the parent latex file can
%% be accessed with the `import' package (which may need to be
%% installed) using
%%   \usepackage{import}
%% in the preamble, and then including the image with
%%   \import{<path to file>}{<filename>.pdf_tex}
%% Alternatively, one can specify
%%   \graphicspath{{<path to file>/}}
%% 
%% For more information, please see info/svg-inkscape on CTAN:
%%   http://tug.ctan.org/tex-archive/info/svg-inkscape
%%
\begingroup%
  \makeatletter%
  \providecommand\color[2][]{%
    \errmessage{(Inkscape) Color is used for the text in Inkscape, but the package 'color.sty' is not loaded}%
    \renewcommand\color[2][]{}%
  }%
  \providecommand\transparent[1]{%
    \errmessage{(Inkscape) Transparency is used (non-zero) for the text in Inkscape, but the package 'transparent.sty' is not loaded}%
    \renewcommand\transparent[1]{}%
  }%
  \providecommand\rotatebox[2]{#2}%
  \newcommand*\fsize{\dimexpr\f@size pt\relax}%
  \newcommand*\lineheight[1]{\fontsize{\fsize}{#1\fsize}\selectfont}%
  \ifx\svgwidth\undefined%
    \setlength{\unitlength}{494.04985563bp}%
    \ifx\svgscale\undefined%
      \relax%
    \else%
      \setlength{\unitlength}{\unitlength * \real{\svgscale}}%
    \fi%
  \else%
    \setlength{\unitlength}{\svgwidth}%
  \fi%
  \global\let\svgwidth\undefined%
  \global\let\svgscale\undefined%
  \makeatother%
  \begin{picture}(1,0.41853571)%
    \lineheight{1}%
    \setlength\tabcolsep{0pt}%
    \put(0,0){\includegraphics[width=\unitlength,page=1]{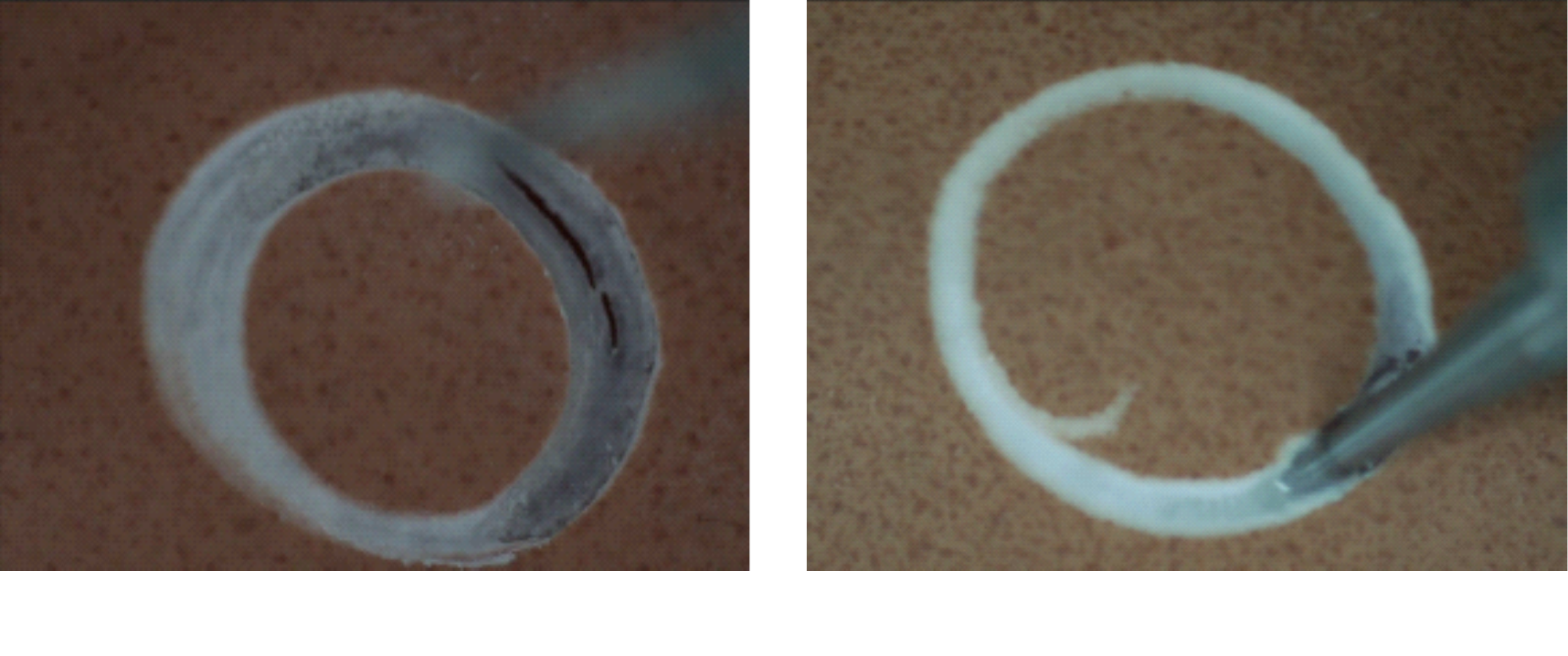}}%
    \put(0.23897472,0.0081438){\color[rgb]{0,0,0}\makebox(0,0)[t]{\lineheight{1.25}\smash{\begin{tabular}[t]{c}(a)\end{tabular}}}}%
    \put(0.7573383,0.0081438){\color[rgb]{0,0,0}\makebox(0,0)[t]{\lineheight{1.25}\smash{\begin{tabular}[t]{c}(b)\end{tabular}}}}%
  \end{picture}%
\endgroup%

\caption{\label{fig:result-drill}Two failure case of the experiments (a) A
failure case caused by tilted initial pose; (b) A failure case caused
by prediction errors of completion level.}
\end{figure}

A tilted initial pose of the egg may also cause drilling failure,
as shown in Fig.$\,$\ref{fig:result-drill}-(a). The failure cases
are due to the fact that the drill bit rubs the membrane surface despite
retaining its z-position at points reaching a 100\% completion level
when the drill is trying to drill the other points that are not. Such
failures accounted for 2 of the 4 total cases of failure. In this
initial proposal, theautonomous drilling process was sensitive to
the initial tilt of the eggs and this is being addressed as part of
future work.

The reason for the other 2 cases of failure is that the areas whose
completion level is already 100 are predicted as less than 100 by
the prediction model, resulting in a continuous drop of the drill
at those areas, rupturing the membrane, as shown in Fig.$\,$\ref{fig:result-drill}-(b).
That refers that we should still refine the neural network for a higher
recognition accuracy while being in real time.

Furthermore, the experiment result also demonstrated the possibility
of generalizing the system for the mouse skull drilling experiments.
It is to be noted that the trajectory planning block of the system
can be promoted directly to the experiment on mice skull while transfer
learning is additionally necessary for the generalizing of the completion
level recognition block since the the training datasets differ.

\section{Conclusion}

In this paper, we proposed a autonomous robotic drilling system for
cranial window creation. To achieve this, a new algorithm for trajectory
planning while avoiding overshoot is imposed. In order to adjust the
generated trajectory in real-time, we applied a neural network to
recognize the completion level of drilling in real-time. In the experiment,
we showed that our robotic drilling system is able to achieve autonomous
drilling at a reasonable success rate and speed with an egg model
and demonstrates the possibility of the system on mice cranial window
creation procedure.

Future works include increasing the drilling speed by improving the
trajectory planner to accommodate for initially tilted eggs, improving
the success rate of drilling by including multimodal information such
as contact force, and the application of the system to drilling mouse
skulls.

\bibliographystyle{IEEEtran}
\bibliography{ICRA2022,IEEEabrv,IEEEexample}

\end{document}